\documentclass{article}
\usepackage[utf8]{inputenc}
\usepackage{graphicx}
\usepackage{amssymb}
\usepackage{url,lineno,microtype}
\usepackage{hyperref}
\usepackage{subcaption}
\usepackage{authblk}
\usepackage{accents}
\usepackage{multirow}
\usepackage{booktabs}
\usepackage{tikzsymbols}

\title{Gaussian Latent Dirichlet Allocation for Discrete Human State Discovery}
\author[1]{Congyu Wu\footnote{Correspondence: congyu.wu@austin.utexas.edu}}
\author[2]{Aaron Fisher}
\author[1]{David Schnyer}
\affil[1]{Department of Psychology, University of Texas, Austin}
\affil[2]{Department of Psychology, University of California, Berkeley}

\date{June 2022}

\begin{document}

\maketitle

\begin{abstract}

In this article we propose and validate an unsupervised probabilistic model, Gaussian Latent Dirichlet Allocation (GLDA), for the problem of discrete state discovery from repeated, multivariate psychophysiological samples collected from multiple, inherently distinct, individuals. Psychology and medical research heavily involves measuring potentially related but individually inconclusive variables from a cohort of participants to derive diagnosis, necessitating clustering analysis. Traditional probabilistic clustering models such as Gaussian Mixture Model (GMM) assume a global mixture of component distributions, which may not be realistic for observations from different patients. The GLDA model borrows the individual-specific mixture structure from a popular topic model Latent Dirichlet Allocation (LDA) in Natural Language Processing and merges it with the Gaussian component distributions of GMM to suit continuous type data. We implemented GLDA using STAN (a probabilistic modeling language) and applied it on two datasets, one containing Ecological Momentary Assessments (EMA) and the other heart measures from electrocardiogram and impedance cardiograph. We found that in both datasets the GLDA-learned class weights achieved significantly higher correlations with clinically assessed depression, anxiety, and stress scores than those produced by the baseline GMM. Our findings demonstrate the advantage of GLDA over conventional finite mixture models for human state discovery from repeated multivariate data, likely due to better characterization of potential underlying between-participant differences. Future work is required to validate the utility of this model on a broader range of applications.

\end{abstract}

\section{Introduction} \label{sec:intro}

In social science and medical research, a prevalent mode of data collection is measuring a set of potentially related variables repeatedly at different points in time from a cohort of distinct subjects. For example, when assessing mental wellbeing, psychologists often use a suite of survey questions to solicit the respondent's in-the-moment experience of multiple mood aspects such as anxiety, sadness, and anger. In the medical context, cardiac, urinary, and serum measures are also extracted simultaneously to form a comprehensive evaluation of a patient's symptoms and potential causes. These repeated multivariate measures are the backbone of both one-time diagnoses and continued monitoring of illnesses, therefore modeling methods for this type of data are important and applicable to a wide variety of domains. Clinical diagnosis entails categorizing a multivariate observation into a \textit{class}, or discrete state, that represents a particular kind of illness or the absence of it. However, at any given time, each variable individually likely does not offer sufficient information to reach such categorization. It is often a combination of values across multiple symptom variables that allows doctors or scientists to confidently determine the type of health or behavioral issue. Due to the limited indicativeness of individual variables and the prevalent codependency between variables to form patterns, clustering multivariate observations into compound states is a crucial operation in clinical data science. Knowing the underlying state of a patient at the moment when a multivariate observation is taken may help healthcare providers correctly diagnose patients without being distracted by possibly spurious changes in individual variables in the sample.

Finite mixture models (FMMs) are a popular class of models used to tackle the problem of clustering momentary multivariate observations into states. Compared to partition-based clustering methods such as k-means, FMMs theorize a soft, probabilistic boundary between classes or components, which is especially fitting for psychophysiological and behavioral data. Because, as opposed to applications such as clustering plants or animals into species, boundaries between human behavioral and health states can be fuzzy. FMMs assume any data point is sampled from a mixture of underlying component distributions, which serve as clusters. The probability that a component distribution is to be sampled from is governed by its corresponding mixture weight. The parameters of such a model, including the mixture weights and the means and covariances of the component distributions, are then inferred iteratively based on data. Due to the shared mixture weights across all data used to fit an FMM, the model is best suited for clustering observations from a purportedly homogeneous source, such as from the same subject. When we collect repeated multivariate samples from a cohort of subjects, using FMM to infer underlying compound states becomes potentially problematic. On the one hand, if we fit an FMM to all data belonging to multiple distinct subjects, we are forcing different subjects' observations to be sampled from one shared mixture, which is a potentially unrealistic assumption and will potentially result in unreliable clusters and weights. On the other hand, if we fit one FMM for each subject, the clusters and weights inferred are for each subject themselves and therefore lacks a mechanism to generalize across and compare between subjects. Current work is lacking \cite{fisher2018lack} in assessing the potential commonality of the states obtained by individually-fitted FMMs. 

To address this limitation, we propose and evaluate a probabilistic model called Gaussian Latent Dirichlet Allocation (GLDA) to combine the advantages of individually-fitted and globally-fitted FMMs and circumvent their respective shortcomings. GLDA was previously and separately introduced in the field of Natural Language Processing as a topic model \cite{das2015gaussian}, serving as a continuous extension of the discrete-input Latent Dirichlet Allocation \cite{blei2003latent}. To the best of our knowledge, no existing work has investigated the applicability and benefit of the GLDA model for clustering multivariate psychophysiological data. As opposed to a probabilistic structure where data points are directly sampled from a global mixture of components, GLDA utilizes a Dirichlet prior to allow different subjects' observations to be sampled from each individual's own weighted mixture while at the same time keeping a global set of components. The ability for each individual to have their own mixture preserves idiographic differences, which is theoretically superior to fitting one FMM to data from inherently distinct subjects. A global, rather than local, set of components is desirable because clinically meaningful symptoms should be generalizable to different individuals. Practically, the global set of components allows observations from different subjects to be compared, which is not straightforwardly achievable by individually-fitted FMMs. Because of the more intricate dependency structure of GLDA, we hypothesize that the class components and the weights inferred by GLDA will be more meaningful, i.e., better predictive of clinically validated individual outcomes, than those obtained by traditional FMMs. 

To test this hypothesis, we built our own implementation of GLDA using STAN \cite{carpenter2017stan}, a probabilistic modeling and inference language, and applied it on two datasets that contain measures of participants' psychology (from ecological momentary assessment or EMA) and physiology (from heart monitors) respectively. Both datasets contain repeated multivariate momentary samples from a cohort of participants and health outcomes clinically diagnosed for each participant. Detailed information about the two datasets is listed in Table \ref{tab:data}. For each dataset, we fit a GLDA model to learn the individualized mixture weights and then used these weights to predict individual health outcomes using linear regression. As a baseline, we built a global Gaussian Mixture Model (GMM, a popular member of FMMs) for each dataset and use the realized class frequencies as class weights of each individual to predict the same health outcomes also using linear regression. We compared the significance of the linear regression models built with class weights learned by GLDA and GMM respectively. 

Our results show that GLDA achieved predominantly greater significance and variance explained for both datasets. This finding indicates that for at least comparable types of cohorts and measures (i.e., psychophysiological), GLDA is indeed a superior clustering method to GMM for discovering states from repeated multivariate samples collected from different subjects. In the remainder of this article, we detail our experiments, present our results, and reflect on the model's implications.

\begin{table}[]
\centering
\normalsize
\begin{tabular}{@{ }lll@{ }}
\toprule
    Model & Plate notation & Sampling process\\ \midrule
    \multirow{9}{*}{GMM} & & Sample mixture weights \\
    & \multirow{9}{*}{\includegraphics[width=0.4\textwidth]{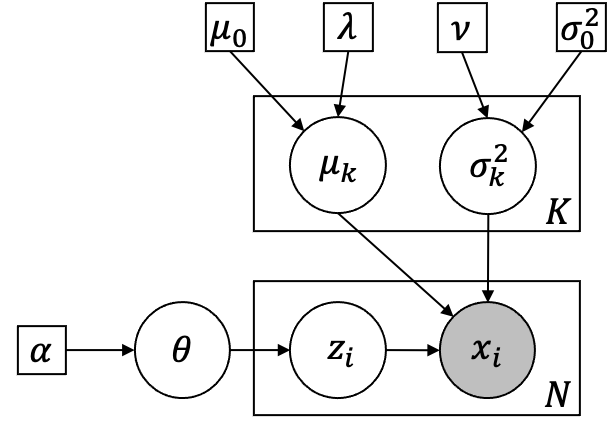}} & \hspace{8pt}$\theta \sim Dirichlet(\alpha)$;  \\ 
    & & For each $k = \{1,...,K\}$:  \\ 
    & & \hspace{8pt}Sample component mean \\
    & & \hspace{16pt}$\mu_k \sim \mathcal{N}(\mu_0, \lambda)$, \\ 
    & & \hspace{8pt}Sample component covariance \\
    & & \hspace{16pt}$\sigma^2_k \sim \mathcal{W}^{-1}(\nu, \sigma^2_0)$; \\
    & & For each $i = \{1,...,N\}$: \\ 
    & & \hspace{8pt}Sample class \\
    & & \hspace{16pt}$z_i \sim Categorical(\theta)$,\\
    & & \hspace{8pt}Sample observation value \\
    & & \hspace{16pt}$x_i \sim \mathcal{N}(\mu_{z_i}, \sigma^2_{z_i})$. \\
\midrule
    \multirow{9}{*}{LDA} & & For each $m = \{1,...,M\}$:\\ 
    & \multirow{9}{*}{\includegraphics[width=0.4\textwidth]{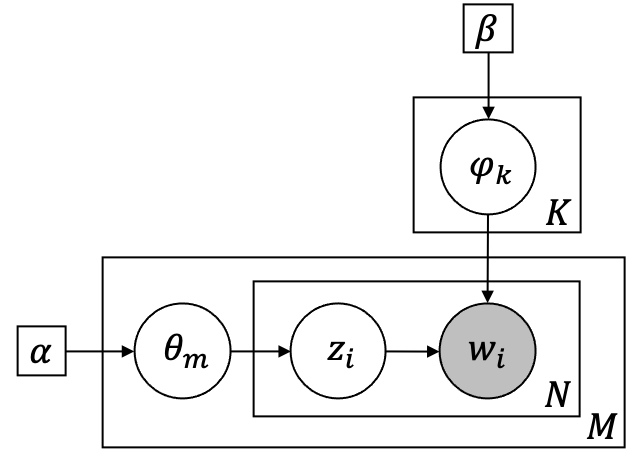}} & \hspace{8pt}Sample topic distribution\\
    & & \hspace{16pt}$\theta_m \sim Dirichlet(\alpha)$; \\ 
    & & For each $k = \{1,...,K\}$:\\ 
    & & \hspace{8pt}Sample word distribution\\
    & & \hspace{16pt}$\phi_k \sim Dirichlet(\beta)$;\\
    & & For each $i = \{1,...,N\}$:\\
    & & \hspace{8pt}Sample topic \\
    & & \hspace{16pt}$z_i \sim Categorical(\theta_m)$, \\
    & & \hspace{8pt}Sample word\\  
    & & \hspace{16pt}$w_i \sim Categorical(\phi_{z_i})$. \\
\midrule
    \multirow{9}{*}{GLDA} & & For each $m = \{1,...,M\}$:\\
    & \multirow{9}{*}{\includegraphics[width=0.4\textwidth]{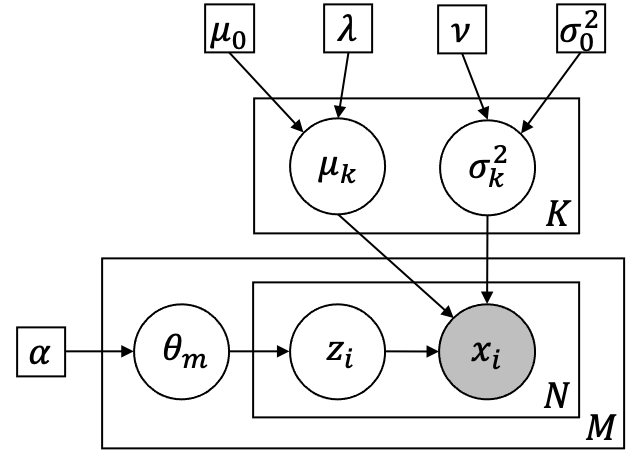}} & \hspace{8pt}Sample mixture weights\\
    & & \hspace{16pt}$\theta_m \sim Dirichlet(\alpha)$; \\ 
    & & For each $k = \{1,...,K\}$:\\
    & & \hspace{8pt}Sample component mean \\
    & & \hspace{16pt}$\mu_k \sim \mathcal{N}(\mu_0, \lambda)$, \\ 
    & & \hspace{8pt}Sample component covariance \\
    & & \hspace{16pt}$\sigma^2_k \sim \mathcal{W}^{-1}(\nu, \sigma^2_0)$; \\
    & & For each $i = \{1,...,N\}$: \\ 
    & & \hspace{8pt}Sample class \\
    & & \hspace{16pt}$z_i \sim Categorical(\theta_m)$,\\
    & & \hspace{8pt}Sample observation value \\
    & & \hspace{16pt}$x_i \sim \mathcal{N}(\mu_{z_i}, \sigma^2_{z_i})$. \\
\bottomrule
\end{tabular}
\caption{Comparison of the statistical dependencies and sampling processes of Gaussian Mixture Model (GMM), Latent Dirichlet Allocation (LDA), and Gaussian Latent Dirichlet Allocation (GLDA). Variables in smaller squares are pre-set hyperparameters while the ones in circles are sampled random variables. Note that GLDA's structure has the top half of GMM and the bottom half of LDA.}\label{tab:glda}
\end{table}

\section{Gaussian Latent Dirichlet Allocation}\label{sec:glda}

The statistical dependency (illustrated with plate notation) and sampling process of GLDA is provided in Table \ref{tab:glda}, together with those of GMM and LDA for a straightforward comparison. The three models are all unsupervised, generative, probabilistic clustering models. All three models require the researcher to supply the number of clusters $K$ to start running (as opposed to Dirichlet Process GMM \cite{gorur2010dirichlet}, a variant of GMM that learns the number of clusters automatically). The GMM shown in Table \ref{tab:glda} is a generalized, Bayesian version of the model; a non-Bayesian GMM would simply be a Bayesian GMM with uninformative priors (hyperparameters). Comparing the plate notations of GMM and GLDA, one should notice that in GMM there is one mixture weight vector $\theta$ for the sampling of all $z_i, i \in \{1,...,N\}$, where $N$ is the total number of observations; whereas in GLDA, a separate mixture weight vector $\theta_m$ is sampled for each participant $m \in \{1,...,M\}$ where $M$ is the total number of participants. Accordingly, the class and value of each observation belonging to participant $m$ would be sampled from the participant-specific mixture weights. Both GMM and GLDA define the component distributions as Gaussian distributions parameterized by mean $\mu_k$ and covariance $\sigma^2_k$, indicated by the identical top half of their plate notations. Because of their Gaussian components, both GMM and GLDA are better suited for continuous type samples. 

LDA, on the other hand, is only applicable to discrete data. It is primarily used for the NLP task of \textit{topic modeling} for a collection of potentially thematically different documents filled with words, since words in human languages are typically considered disjoint units. A \textit{topic} is simply a weight vector $\phi$ over a comprehensive collection of words, or the \textit{vocabulary}, that sum up to 1. Intuitively, the semantic make-up of a topic is determined by the relative weights over different words. LDA assumes that for each document $m \in \{1,...,M\}$ there is a specific mixture of $K$ topics (topic distribution), governed by a mixture weight vector $\theta_m$. To sample a word $w_i$ in document $m$, LDA first samples a topic $z_i \in \{1,...,K\}$ according to the document-specific topic distribution $\theta_m$, then samples a word according to the topic-specific word distribution $\phi_{z_i}$. Such a document-topic-word hierarchy corresponds straightforwardly to the participant-class-observation hierarchy in GLDA. Note the identfical structure of the bottom half of LDA's and GLDA's plate notation: just as each document has a separate topic distribution in LDA, each participant has a separate set of mixture weights in GLDA. The difference is in how the words/observations are sampled after a topic/class $z_i$ is sampled: in LDA a word is sampled from a categorical distribution parameterized by $\phi_{z_i}$ whereas in GLDA, each observation is sampled from a Gaussian distribution parameterized by $\mu_{z_i}$ and $\sigma^2_{z_i}$ (as the initial ``G" suggests). Marrying the characteristics of GMM and LDA, GLDA allows the sampling of continuous values from participant-specific mixture weights over discrete states. We suspect this is an attractive structure for human state discovery from repeated multivariate psychophysiological samples. 

\begin{table}[]
\begin{tabular}{@{ }lp{0.4\textwidth}p{0.35\textwidth}@{ }}
\toprule
                   & Dataset 1 (Psychology) & Dataset 2 (Physiology) \\  \midrule
           \textbf{Subjects} & & \\ \midrule
           Source  & Clinical patients & Clinical patients  \\
           Sex     & 69\% female  & 61\% female      \\ 
           Age     & Mean age 38.5 years  & Mean age 39 years \\  
           Number $M$ & 45    & 166  \\ \midrule
           \textbf{Samples} && \\ \midrule
         Instrument & EMA   & MindWare Mobile Hardware  \\ 
         Variables  & Self-report on 0-100 (4 times/day) of: rumination, worry, fear, anger, irritability, anhedonia, hopelessness, depressed mood, avoidance of action, avoidance of people    & Sensor reading (500 Hz) of: heart rate, respiratory sinus arrhythmia, pre-ejection period, respiration rate, respiration amplitude           \\ 
         Dimension $V$  & 10     & 5     \\
         Total $N$    & 5,076     & 25,394    \\ \midrule
\textbf{Ground truth} &  & \\ \midrule
        Depression & \multicolumn{2}{l}{HRSD score (Hamilton Rating Scale for Depression, 52 pts)} \\  
        Anxiety & \multicolumn{2}{l}{HAMA score (Hamilton Anxiety Rating Scale, 56 pts)} \\  
        Stress & \multicolumn{2}{l}{DASS-S score (Depression Anxiety and Stress Scale--Stress, 42 pts)} \\  
\bottomrule
\end{tabular}
\caption{Summary of the two datasets}
\label{tab:data}
\end{table}

\section{Data}

As mentioned in Section \ref{sec:intro}, we use two datasets to validate GLDA. Dataset 1 \cite{fisher2020identifying} contains a total of 5,076 time-stamped Ecological Momentary Assessment (EMA) samples of ten mental health items from a group of 45 participants (69\% female; mean age 38.5 years, s.d. 13.8 years). The participants were asked to self-report four times a day for at least 30 days consecutively their momentary experience of (1) rumination, (2) worry, (3) fear, (4) anger, (5) irritability, (6) anhedonia, (7) hopelessness, (8) depressed mood, (9) avoidance of action, and (10) avoidance of people on a scale of 0-100. In addition, participants’ \textit{trait} depression, anxiety, and stress were diagnosed pre-study using the following clinical inventories: for depression, the Hamilton Rating Scale for Depression (HRSD, 17 questions, 0-52 points) \cite{hamilton1960rating}; for anxiety, the Hamilton Anxiety Rating Scale (HAMA, 14 questions, 0-56 points) \cite{hamilton1959assessment}, and; for stress, the stress subscale from the Depression Anxiety and Stress Scale (DASS-S, 14 questions each rated on 0, 1, 2, and 3 with a total score of 42 for each outcome) \cite{lovibond1995structure}. The DASS inventory also contains a depression (DASS-D) and an anxiety (DASS-A) subscale; however, these were eschewed in favor of the HRSD and HAMA scores as the official ground truth for depression and anxiety. 

Dataset 2 \cite{fisher2022unsupervised} contains 25,394 physiological samples of 5 physiological measures from 166 participants (61\% female; mean age 39 years, s.d. 13.8 years), a superset of the group of 45 in Dataset 1. The same depression, anxiety, and stress diagnosis scores from HRDS, HAMA, and DASS-S are available for each of the 166 participants as well. Continuous electrocardiogram (ECG) and impedance cardiography data were collected at 500 Hz using the MindWare Mobile Hardware during a clinical interview of each participant. From these data we extracted five physiological variables for use in this study: (1) heart rate (hr), (2) respiratory sinus arrhythmia (rsa), (3) pre-ejection period (pep), (4) respiration rate (resp\_rate), and (5) respiration amplitude (resp\_amp). Each of these five variables forms a time series. 

The details described above are summarized in Table \ref{tab:data}. The commonality between, as well as the reasons we choose these two datasets, are three fold. First, they fit the type of data in question in this article: repeated, multivariate samples from a cohort of distinct subjects. Second, inferring discrete states from these ``atomic" variables is a meaningful practice for automated diagnostic purposes \cite{fisher2020identifying}. Last but not least, ground truth about each participant is available, in both cases trait depression, anxiety, and stress, allowing the validation of the model-learned mixtures of discrete states against the participants' actual clinical outcomes.

\section{Experiments}\label{sec:experiments}

We used STAN, a programming language for probabilistic modeling and inference to specify our GLDA models. We found one existing implementation of GLDA (in Python) for topic modeling \url{https://pypi.org/project/gaussianlda/}, taking in words as input and involving the conversion of words into real numbers using a word-embedding application. We created our own implementation to directly process continuous data. The STAN code is provided in the Supplementary Material. The number of $K$, the number of discrete states, or clusters, is the only parameter that needs to be set by the researcher. Hyperparameters were initiated by the STAN program automatically to have neutral or unit values. The rest of the input parameters are directly read from the data (see Table \ref{tab:data}), which include: (1) $V$, the number of variables, or dimension, in the multivariate samples; (2) $M$, the number of distinct participants in the cohort, and; (3) $N$, the total number of observations. 

The next step is fitting the model to data. We standardized each variable within each participant so that they follow a standard normal distribution (Z-score; 0 mean and 1 standard deviation). This operation preserves the interrelations between variables but ensures that the discrete states discovered by our model reflect momentary deviations from intra-individual mean values. For a simple, demonstrative clustering model, we set $K=3$, and ran the STAN program to find the participant-specific mixture weights $\theta_m$ (each $\theta_m$ is a $K$-length weight vector that sums up to 1), as well as the mean $\mu_k$ ($V$-length vector) and covariance matrix $\sigma^2_k$ ($V*V$ matrix) of each cluster $k$. The mean and covariance effectively defined each cluster as a distribution over the multiple variables. We configured the STAN program to run a 1000-iteration Markov Chain Monte Carlo chain including an initial 200 warm-up iterations and calculated the mean values of the output parameters from all the iterations following the warm-up. Based on our observation, it safely takes $<$100 iterations to warm up so convergence has long been reached when the sampling algorithm stops running and results are returned.

We repeated the process described above for both datasets. We then explored the correlation between the participants' clinically diagnosed depression, anxiety, and stress levels with the mixture weights inferred by GLDA on the EMA data (Dataset 1) and physiological data (Dataset 2). We built linear regression models of HRSD depression score, HAMA anxiety score, DASS stress score, using the GLDA-learned participant-specific mixture weights. In doing so we aimed to validate our GLDA-discovered discrete states because intuitively they should be significantly correlated with the results from the established inventories. Returning significant coefficients would support the validity of our inferred clusters and their weights. A lack of correlation between machine learned discrete states from EMAs and the one-time assessment from inventories would indicate that the learned states are not capturing significant variance in the participants' mental health status. A strong correlation, on the other hand, would validate the discovered classes and their individual-specific weights. 

Finally, to determine whether and to what extent the GLDA-discovered classes may be more meaningful than regular GMM, we built a 3-class GMM on all observations in each dataset, producing one set of mixture weights over component distributions. We computed the class membership of each observation and then the class proportions for each participant as the frequency of each class in each participant divided by the number of observations from the participant. Linear regression models of HRSD, HAMA, and DASS-S scores were then built using the individual-specific class proportions as predictors. We compared the results of these GMM-based linear models with those of the GLDA-based ones to identify which method may produce more meaningful or predictive classes and weights.  

\section{Results}\label{sec:results}

\begin{figure}[]
  \begin{subfigure}[b]{0.48\textwidth}
         \centering
         \includegraphics[width=\textwidth]{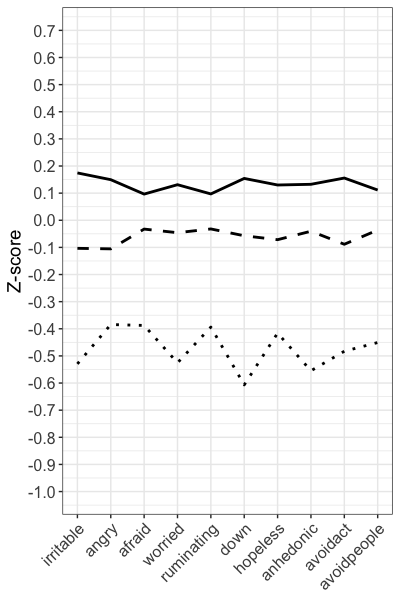}
         \caption{GLDA cluster mean $\mu_k, k = \{1,2,3\}$}
         \label{fig:glda_mu_dataset1}
     \end{subfigure}
     \begin{subfigure}[b]{0.48\textwidth}
         \centering
         \includegraphics[width=\textwidth]{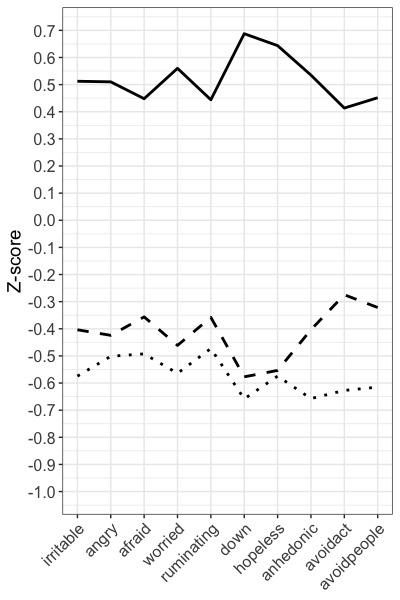}
         \caption{GMM cluster mean $\mu_k, k = \{1,2,3\}$}
         \label{fig:gmm_mu_dataset2}
     \end{subfigure}
 \caption{Mean values of the three clusters (solid, dashed, and dotted lines) over the ten, momentarily evaluated mood items in Dataset 1, learned by GLDA and GMM.}
 \label{fig:mu_dataset1}
\end{figure}

\subsection{Dataset 1}\label{subsec:dataset1}

Figure \ref{fig:mu_dataset1} shows the GLDA- and GMM-inferred clusters' means over the ten, momentarily evaluated mood items. The three classes learned by both methods correspond to an unwell state (solid line; high Z-scores across all variables), a well state (dotted line), and a middle state (dashed line). The middle state from both GLDA and GMM stays on the negative side over all ten variables, indicating a mildly positive experience. No clusters appear to contain low and high values in different variables simultaneously (i.e., lines not crossing), likely due to the negative valence of all the EMA items evaluated. Two differences between the cluster means are visible. First, the unwell state by GMM is more severe than its GLDA counterpart by a significant margin. Second, in the GMM result, outside of the separation observed in \textit{avoidance of action} and \textit{avoidance of people}, the mean values of the middle state nearly overlap with those of the well state; whereas the middle state defined by GLDA exhibits better separation from both the well and the unwell states. Overall, the cluster means from the two methods are more similar than they are different. 

\begin{table}
\centering
\begin{tabular}{@{ }llp{0.1\textwidth}p{0.1\textwidth}p{0.1\textwidth}p{0.1\textwidth}p{0.1\textwidth}p{0.1\textwidth}@{ }}
\toprule
 & & \multicolumn{2}{c}{HRSD (Depression)} & \multicolumn{2}{c}{HAMA (Anxiety)} & \multicolumn{2}{c}{DASS-S (Stress)}\\ \cmidrule{3-8}
 & & $p$ & $adj.r^2$ & $p$ & $adj.r^2$ & $p$ & $adj.r^2$ \\  
\midrule
\multirow{3}{*}{GLDA} & \Xey[1.2] &  \textbf{0.024}* & 9.2\% & \textbf{0.010}* & 12.5\% & \textbf{0.005}** & 15.1\% \\ \cmidrule{2-8}
     & \Smiley[1.2] & 0.055. & 6.2\% & 0.085. & 4.6\% & 0.140 & 2.8\% \\ \cmidrule{2-8}
     & \Laughey[1.2] & 0.815 & -2.2\% & 0.348 & -0.2\% & 0.107 & 3.7\% \\     
\midrule
\multirow{3}{*}{GMM} &  \Xey[1.2] & 0.655 & -1.8\% & 0.884 & -2.3\% & 0.068. & 5.4\% \\ \cmidrule{2-8}
    & \Smiley[1.2]& 0.677 & -1.9\% & 0.055. & 6.1\% & 0.160 & 2.3\% \\ \cmidrule{2-8}
    & \Laughey[1.2] & 0.955 & -2.3\% & 0.161 & 2.3\% & 0.051. & 6.5\% \\ 
\bottomrule
\end{tabular}
\caption{Results of individual-level linear regression models of clinically evaluated depression, anxiety, and stress, using mixture weights of the three classes learned by GLDA and the realized class proportions learned by GMM. Emojis \Xey[1.2], \Smiley[1.2], and \Laughey[1.2] represent the unwell, middle, and well states visualized in Figure \ref{fig:mu_dataset1} respectively. Significant pairs of class weight/proportion and mental illness score are shown in bold and marked according to its significance level (**: $0.001\leq p<0.01$; *: $0.01\leq p<0.05$; .: $0.05\leq p<0.1$)}
\label{tab:glda_correlations1}      
\end{table}

In Table \ref{tab:glda_correlations1}, we list the p-values and adjusted R squared values of the linear regression models predicting HRSD, HAMA, and DASS-S scores with individualized mixture weights by GLDA and realized class proportions by GMM from the three-class clustering analysis described in Section \ref{sec:experiments}. Each set of p-values and adjusted R squared values come from a univariate linear regression between a pair of clustering weight/proportion and a ground truth variable. Evidently, the only significant pairs (at 0.05 level of confidence) are between the weight of the GLDA-learned unwell state (\Xey[1.2]) and each of the clinical scores. The strongest correlation is between the GLDA-learned unwell state weight and DASS-S stress score: the former accounted for 15.1\% of the variance in the latter. The GMM-learned unwell state proportion is not significantly correlated with clinical scores, neither are the weights/proportions of the middle and the well states (\Smiley[1.2] and \Laughey[1.2]) from either methods. This result supports the hypothesis that the class components and the weights inferred by GLDA are more meaningful, i.e., better predictive of clinically validated individual outcomes, than those obtained by GMM. This result also conforms to the intuition that the unwell state should be correlated with clinical ground truth of mental illness due to the shared underpinning of negative affect.

\begin{figure}[]
  \begin{subfigure}[b]{0.48\textwidth}
         \centering
         \includegraphics[width=\textwidth]{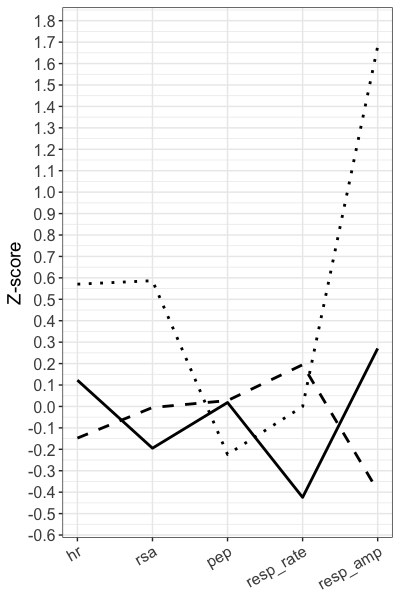}
         \caption{GLDA cluster mean $\mu_k, k = \{1,2,3\}$}
         \label{fig:glda_mu_dataset2}
     \end{subfigure}
     \begin{subfigure}[b]{0.48\textwidth}
         \centering
         \includegraphics[width=\textwidth]{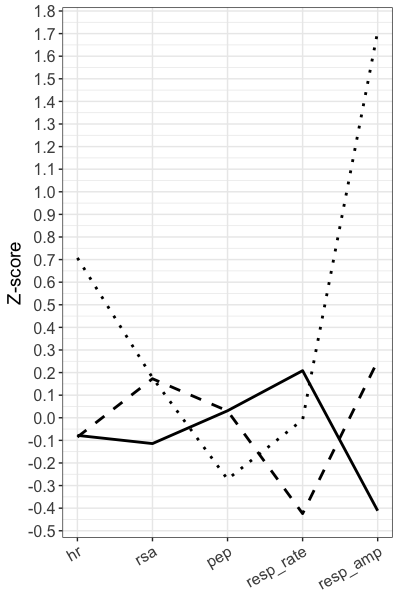}
         \caption{GMM cluster mean $\mu_k, k = \{1,2,3\}$}
         \label{fig:gmm_mu_dataset2}
     \end{subfigure}
 \caption{Mean values of the three clusters (solid, dashed, and dotted lines) over the five physiological sensor measures in Dataset 2, learned by GLDA and GMM. The five measures, from left to right on the X-axis, are heart rate (hr), respiratory sinus arrhythmia (rsa), pre-ejection period (pep), respiration rate (resp\_rate), and respiration amplitude (resp\_amp).}
 \label{fig:mu_dataset2}
\end{figure}

\subsection{Dataset 2}\label{subsec:dataset2}

Figure \ref{fig:mu_dataset2} shows the GLDA- and GMM-inferred clusters' means over the five physiological measures from MindWare Mobile Hardware. The three classes discovered by both methods again have similar shapes but also exhibit key differences. There is a conspicuous high heart rate (hr) and high respiratory amplitude (resp\_amp) class from both methods (dotted line); but GLDA associates high respiratory sinus arrhythmia (rsa) with the state whereas GMM does not. The other two, relatively lower arousal states (solid and dashed lines) from GLDA and GMM differ in the combination of high and low values in all variables except pre-ejection period (pep), whose Z-score is right at zero for both methods.   

\begin{table}
\centering
\begin{tabular}{@{ }lp{0.1\textwidth}p{0.1\textwidth}p{0.1\textwidth}p{0.1\textwidth}p{0.1\textwidth}p{0.1\textwidth}@{ }}
\toprule
 & \multicolumn{2}{c}{HRSD (Depression)} & \multicolumn{2}{c}{HAMA (Anxiety)} & \multicolumn{2}{c}{DASS-S (Stress)}\\ \cmidrule{2-7}
 & $p$ & $adj.r^2$ & $p$ & $adj.r^2$ & $p$ & $adj.r^2$ \\  
\midrule
GLDA high-hr &  \textbf{0.012}* & 3.4\% & \textbf{0.047}* & 1.9\% & \textbf{0.001}** & 5.8\% \\
\midrule
GMM high-hr &  0.107 & 1.0\% & 0.260 & 0.2\% & \textbf{0.015}* & 2.9\% \\ 
\bottomrule
\end{tabular}
\caption{Results of individual-level linear regression models of clinically evaluated depression, anxiety, and stress, using the GLDA-learned mixture weight and the GMM-learned realized proportion of the high heart rate (hr), high respiration amplitude (resp\_amp) state, visualized by dotted line in Figure \ref{fig:mu_dataset2}.}
\label{tab:glda_correlations2}
\end{table}

In Table \ref{tab:glda_correlations2}, we list the p-values and adjusted R squared values of linear regression models predicting HRSD, HAMA, and DASS-S scores with the GLDA-learned mixture weight and the GMM-learned realized proportion of the high heart rate (hr), high respiration amplitude (resp\_amp) state (dotted line in Figure \ref{fig:mu_dataset2}). The weights/proportions associated with the other two classes (solid and dashed lines in Figure \ref{fig:mu_dataset2}) did not return any significant result with clinical mental health outcomes, so we do not include them in Table \ref{tab:glda_correlations2}. Once again, the GLDA-learned mixture weight achieved significant correlation with all three clinical scores, with DASS-S stress being the strongest. Proportion of the GMM-inferred high-hr class was also significantly correlated with DASS-S, but not with depression and anxiety. Between the two correlations with DASS-S, GLDA provided better statistical prediction than GMM (GLDA $p=0.001$ vs. GMM $p=0.015$; the second to last column of Table \ref{tab:glda_correlations2}). This result once again supports the hypothesis that the GLDA-inferred discrete states and their mixture weights are more meaningful compared to the realized class proportions learned by GMM. This result also suggests that high heart rate and high respiratory amplitude are two physiological measures that are robustly predictive of depression, anxiety, and especially stress.

\section{Discussion}\label{sec:discussion}

In this section we reflect on two implications of the GLDA model, each lending itself to follow-up research questions and future work. The first reflection is on the conditions under which GLDA may be a superior model to GMM (as demonstrated in this article) and the conditions under which it may not be the case. What allowed GLDA to discover discrete states from momentary psychophysiological measures whose mixture weights are more significantly correlated with clinical mental health outcomes in this paper? Our hypothesis is that it is because of the inherent psychophysiological differences between the participants in the two datasets we used. It is indeed likely that negative affect (Dataset 1) and cardiovascular health (Dataset 2) are experienced differently by different individuals, allowing the individually separate mixture weight sampling mechanism of GLDA (see Table \ref{tab:glda}) to better reflect this reality. Contrarily, if there are no purportedly intrinsic between-participant differences, then GLDA's additional assumption would be unnecessary and it would ostensibly achieve results that are no better or worse than GMM. Given the higher model complexity in GLDA, in that case using GMM may be the more desirable (economical) choice. Future work is needed to validate GLDA on repeated multivariate samples that measure largely homogeneous processes between participants in a cohort and compare the result with GMM. Such work will further pinpoint the conditions under which GLDA is a more advisable approach compared to regular finite mixture models. 

\begin{figure}[]
 \centering
 \includegraphics[width=1\textwidth]{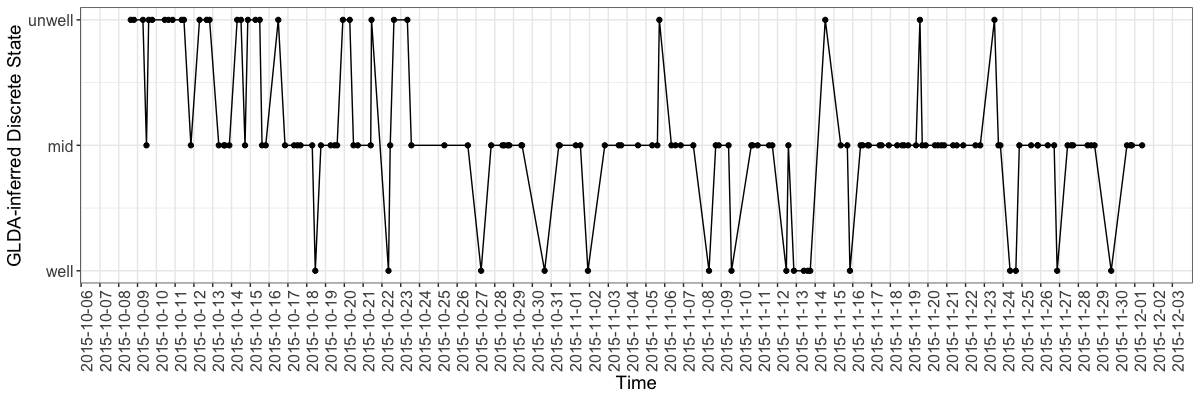}
 \caption{Cluster membership of each timestamped EMA self-report from an example participant (participant P113) in Dataset 1, learned by GLDA. The black dots reflect the moments when a EMA self-report was submitted.}
 \label{fig:glda_cls_26}
\end{figure}

The second reflection is about using the GLDA model to infer not only individual level characteristics (such as mixture weights focused on in this article) but also moment-to-moment transition between discrete states within each participant's observations. In addition to calculating the individual-level mixture weights over clusters, we can also compute the probability that each observation from each participant belongs to each of the $K$ clusters using participant-specific mixture weights $\theta_m$, means $\mu_k$, and covariance matrices $\sigma^2_k$. The formula for this probability is: 

\begin{equation}
    P(z_{n,k}=1|\textbf{x}_n) = \frac{\theta_{m,k}\mathcal{N}(\textbf{x}_n|\mu_k, \Sigma_k)}{\sum_{j=1}^K \theta_{m,j}\mathcal{N}(\textbf{x}_n|\mu_j, \Sigma_j)},
\end{equation}

\noindent where $z_{n,k}$ is a binary variable that takes value 1 if the $n$-th observation $\textbf{x}_n$ (which belongs to participant $m$) belongs to cluster $k$ and 0 otherwise. $\mathcal{N}(\textbf{x}_n|\mu_k, \Sigma_k)$ represents the density value of the multivariate Gaussian distribution $\mathcal{N}(\mu_k, \Sigma_k)$ at $\textbf{x}_n$. The output from this step $P(z_{n,k}=1|\textbf{x}_n)$ where $n=1,2,...,N$, provides observation-level, or momentary insight. Intuitively, cluster $argmax_{k}P(z_{n,k}=1|\textbf{x}_n)$ is assigned the cluster to which observation $\textbf{x}_n$ belongs. The class membership of each observation, which is timestamped, reveals how each participant transitions between different discrete states as time progresses. Figure \ref{fig:glda_cls_26} shows such a time series for an example participant in Dataset 1. One can see that most of the moments that fall in the unwell state gather around an earlier part (2015-10-08 to 2015-10-16) of the study period. The further research question we may ask here is whether those momentary unwell states indicate any temporary behavioral change or health variation? Ground truth for such momentary fluctuation is not available in the datasets used in this article but we defer to future work to validate the utility of these momentary states for predictive, real-time monitoring. Increasingly available, dense, and multi-modal passive sensing data from mobile, wearable, and environmental sensing devices \cite{wu2021multi} should provide plenty of evidence for in-the-moment objective ground truth of an individual's daily behavior and health.

\section{Conclusion}\label{sec:conclusion}

In this article we proposed and validated an unsupervised probabilistic model, Gaussian Latent Dirichlet Allocation (GLDA), for the problem of discrete state discovery from repeated, multivariate psychophysiological samples collected from multiple, inherently distinct, individuals. Psychology and medical research heavily involves measuring potentially related but individually inconclusive variables from cohorts of participants to derive diagnosis, necessitating clustering analysis. Traditional probabilistic clustering models such as Gaussian Mixture Model (GMM) assume a global mixture of component distributions, which may not be realistic for observations from different patients. The GLDA model borrows the individual-specific mixture structure from a popular topic model Latent Dirichlet Allocation (LDA) in Natural Language Processing and merges it with the Gaussian component distributions of GMM to suit continuous type data. 

We implemented GLDA using STAN (a probabilistic modeling language) and applied it on two datasets, one containing Ecological Momentary Assessments (EMA) and the other heart measures from electrocardiogram and imped-ance cardiograph. Both datasets contained clinically diagnosed depression (Hamilton Rating Scale for Depression), anxiety (Hamilton Anxiety Rating Scale), and stress scores (Depression Anxiety and Stress Scale--Stress) for each participant studied. We found that, in both datasets, the GLDA-learned class weights achieved significantly higher correlations with the clinical mental health outcomes than the class proportions discovered by GMM (Sections \ref{subsec:dataset1} and \ref{subsec:dataset2}). Our findings point to an advantage of GLDA over conventional finite mixture models for human state discovery from repeated multivariate data in psychology and medical research, likely due to better characterization of potential underlying between-participant differences. 

Future work is required to validate the utility of this model on a broader range of applications. The purpose of this planned work is twofold, as discussed in Section \ref{sec:discussion}: to elucidate the conditions under which GLDA is and is not an ideal choice for human state discovery from repeated multivariate cohort samples, and to extend the analyses presented in this article from individual-level to momentary, observation-level for real-time health and behavior informatics. Moreover, since the main purpose of this study is to validate GLDA against regular GMM, we chose a singular number of clusters $K=3$ to be simplistic yet realistic; future work can be dedicated to evaluating the sensitivity of model fit and predictive power to different choices of number of clusters.  

\section*{Funding}
This work was supported by Whole Communities—Whole Health, a research grand challenge at the University of Texas at Austin.

\bibliographystyle{plain}
\bibliography{mybib}

\end{document}